\documentclass[10pt,twocolumn,letterpaper]{article}

\usepackage{iccv}
\usepackage{times}
\usepackage{epsfig}
\usepackage{booktabs}
\usepackage{graphicx}
\usepackage{amsmath}
\usepackage{amssymb}
\usepackage{cite}
\usepackage{enumitem}
\usepackage{multirow}

\usepackage{chngcntr}
\usepackage[accsupp]{axessibility} 

\PassOptionsToPackage{hyphens}{url}
\usepackage[pagebackref=true, breaklinks=true, letterpaper=true, colorlinks, bookmarks=false]{hyperref}

\iccvfinalcopy 

\def\iccvPaperID{9889} 

\begin{document}

\title{GameFormer: Game-theoretic Modeling and Learning of Transformer-based Interactive Prediction and Planning for Autonomous Driving}

\author{Zhiyu Huang$^{\dagger}$, Haochen Liu$^{\dagger}$, Chen Lv$^{*}$ \\
Nanyang Technological University, Singapore \\
{\small $^{\dagger}$ Equal contribution {\tt \{zhiyu001,haochen002\}@e.ntu.edu.sg}} \\ 
{\small $^{*}$ Corresponding author {\tt lyuchen@ntu.edu.sg}}
}

\maketitle
\ificcvfinal\thispagestyle{empty}\fi

\begin{abstract}
Autonomous vehicles operating in complex real-world environments require accurate predictions of interactive behaviors between traffic participants. This paper tackles the interaction prediction problem by formulating it with hierarchical game theory and proposing the GameFormer model for its implementation. The model incorporates a Transformer encoder, which effectively models the relationships between scene elements, alongside a novel hierarchical Transformer decoder structure. At each decoding level, the decoder utilizes the prediction outcomes from the previous level, in addition to the shared environmental context, to iteratively refine the interaction process. Moreover, we propose a learning process that regulates an agent's behavior at the current level to respond to other agents' behaviors from the preceding level. Through comprehensive experiments on large-scale real-world driving datasets, we demonstrate the state-of-the-art accuracy of our model on the Waymo interaction prediction task. Additionally, we validate the model's capacity to jointly reason about the motion plan of the ego agent and the behaviors of multiple agents in both open-loop and closed-loop planning tests, outperforming various baseline methods. Furthermore, we evaluate the efficacy of our model on the nuPlan planning benchmark, where it achieves leading performance. Project website: {\small \url{https://mczhi.github.io/GameFormer/}}
\end{abstract}

\section{Introduction}
\label{sec:intro}

\begin{figure}[htp]
    \centering
    \includegraphics[width=\linewidth]{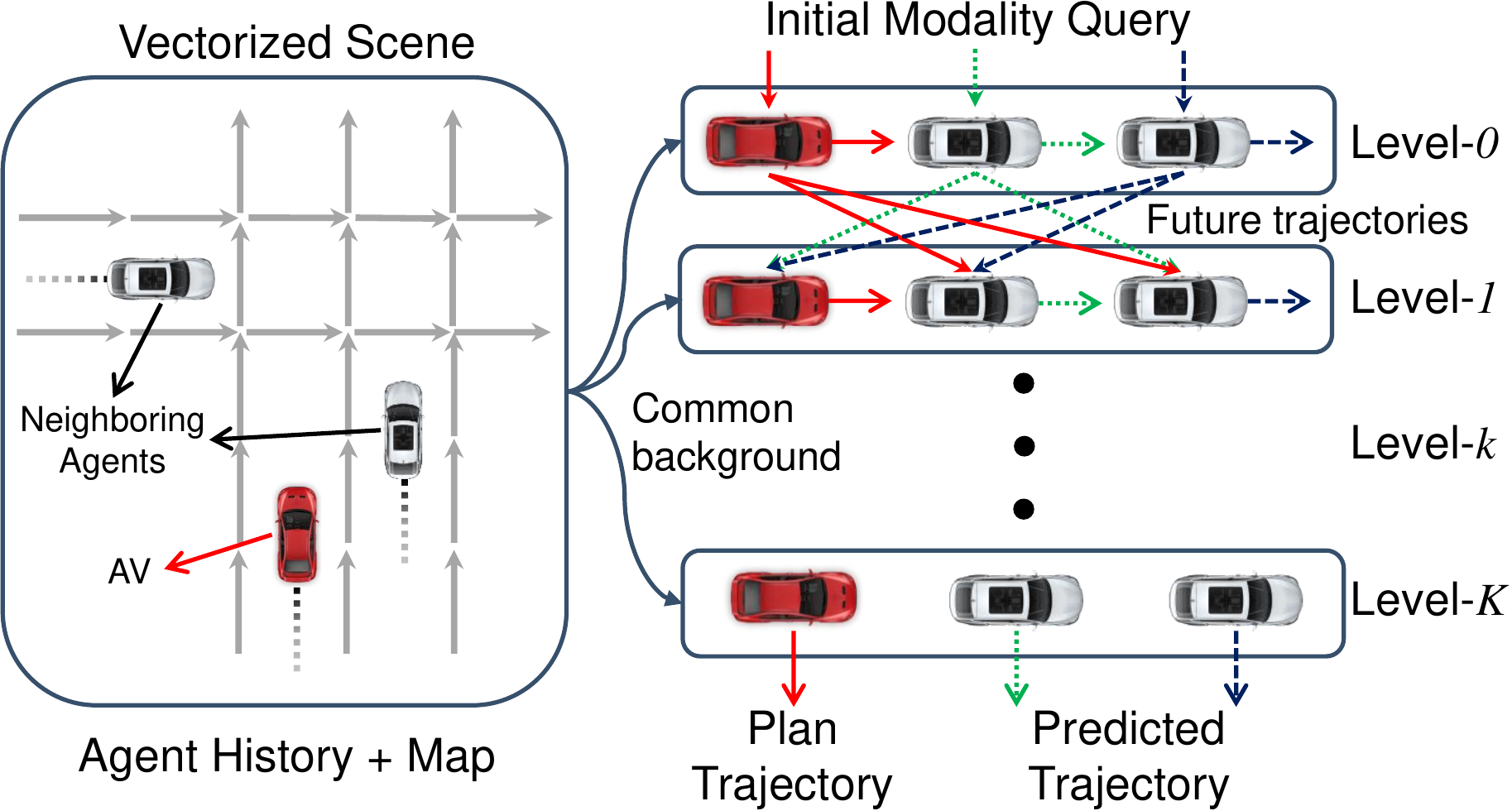}
    \caption{Hierarchical game theoretic modeling of agent interactions. The historical states of agents and maps are encoded as background information; a level-$0$ agent's future is predicted independently based on the initial modality query; a level-$k$ agent responds to all other level-$(k-1)$ agents.}
    \vspace{-0.4cm}
    \label{fig:1}
\end{figure}

Accurately predicting the future behaviors of surrounding traffic participants and making safe and socially-compatible decisions are crucial for modern autonomous driving systems. However, this task is highly challenging due to the complexities arising from road structures, traffic norms, and interactions among road users \cite{gilles2022thomas, jia2022multi, jia2021ide}. In recent years, deep neural network-based approaches have shown remarkable advancements in prediction accuracy and scalability \cite{cui2019multimodal, gao2020vectornet, gu2021densetnt, varadarajan2022multipath++, jia2023towards}. In particular, Transformers have gained prominence in motion prediction \cite{ngiam2021scene, nayakanti2022wayformer, shi2022motion, yuan2021agentformer, zhou2022hivt, jia2022hdgt} because of their flexibility and effectiveness in processing heterogeneous information from the driving scene, as well as their ability to capture interrelationships among the scene elements.

Despite the success of existing prediction models in encoding the driving scene and representing interactions through agents' past trajectories, they often fail to explicitly model agents' future interactions and their interaction with the autonomous vehicle (AV). This limitation results in a passive reaction from the AV's planning module to the prediction results. However, in critical situations such as merge, lane change, and unprotected left turn, the AV needs to proactively coordinate with other agents. Therefore, joint prediction and planning are necessary for achieving more interactive and human-like decision-making. To address this, a typical approach is the recently-proposed conditional prediction model \cite{song2020pip, tolstaya2021identifying, salzmann2020trajectron++, sun2022m2i, huang2022conditional}, which utilizes the AV's internal plans to forecast other agents' responses to the AV. Although the conditional prediction model mitigates the interaction issue, such a one-way interaction still neglects the dynamic mutual influences between the AV and other road users. From a game theory perspective, the current prediction/planning models can be regarded as {\em leader-follower games} with limited levels of interaction among agents.

In this study, we utilize a hierarchical game-theoretic framework (level-$k$ game theory) \cite{costa2009comparing, wright2010beyond} to model the interactions among various agents \cite{li2017game, wang2022social, 8430842} and introduce a novel Transformer-based prediction model named {\em GameFormer}. Stemming from insights in cognitive science, level-$k$ game theory offers a structured approach to modeling interactions among agents. At its core, the theory introduces a hierarchy of reasoning depths denoted by $k$. A level-$0$ agent acts independently without considering the possible actions of other agents. As we move up the hierarchy, a level-$1$ agent considers interactions by assuming that other agents are level-$0$ and predicts their actions accordingly. This process continues iteratively, where a level-$k$ agent predicts others' actions assuming they are level-$(k-1)$ and responds based on these predictions. Our model aligns with the spirit of level-$k$ game theory by considering agents' reasoning levels and explicit interactions.

As illustrated in Fig. \ref{fig:1}, we initially encode the driving scene into background information, encompassing vectorized maps and observed agent states, using Transformer encoders. In the future decoding stage, we follow the level-$k$ game theory to design the structure. Concretely, we set up a series of Transformer decoders to implement level-$k$ reasoning. The level-$0$ decoder employs only the initial modality query and encoded scene context as key and value to predict the agent's multi-modal future trajectories. Then, at each iteration $k$, the level-$k$ decoder takes as input the predicted trajectories from the level-$(k\! - \! 1)$ decoder, along with the background information, to predict the agent's trajectories at the current level. Moreover, we design a learning process that regulates the agents' trajectories to respond to the trajectories of other agents from the previous level while also staying close to human driving data. The main contributions of this paper are summarized as follows:

\begin{enumerate}[topsep=1pt, itemsep=1pt, partopsep=1pt, parsep=1pt]
\item We propose {\em GameFormer}, a Transformer-based interactive prediction and planning framework. The model employs a hierarchical decoding structure to capture agent interactions, iteratively refine predictions, and is trained based on the level-$k$ game formalism.
\item We demonstrate the state-of-the-art prediction performance of our {\em GameFormer} model on the Waymo interaction prediction benchmark.
\item We validate the planning performance of the {\em GameFormer} framework in open-loop driving scenes and closed-loop simulations using the Waymo open motion dataset and the nuPlan planning benchmark.
\end{enumerate}

\section{Related Work}
\subsection{Motion Prediction for Autonomous Driving}
Neural network models have demonstrated remarkable effectiveness in motion prediction by encoding contextual scene information. Early studies utilize long short-term memory (LSTM) networks \cite{alahi2016social} to encode the agent's past states and convolutional neural networks (CNNs) to process the rasterized image of the scene \cite{cui2019multimodal, huang2022recoat, salzmann2020trajectron++, gilles2021home}. To model the interaction between agents, graph neural networks (GNNs) \cite{mo2022multi, huang2022multi, chen2022scept, gilles2022gohome} are widely used for representing agent interactions via scene or interaction graphs. More recently, the unified Transformer encoder-decoder structure for motion prediction has gained popularity, \eg, SceneTransformer \cite{ngiam2021scene} and WayFormer \cite{nayakanti2022wayformer}, due to their compact model description and superior performance. However, most Transformer-based prediction models focus on the encoding part, with less emphasis on the decoding part. Motion Transformer \cite{shi2022motion} addresses this limitation by proposing a well-designed decoding stage that leverages iterative local motion refinement to enhance prediction accuracy. Inspired by iterative refinement and hierarchical game theory, our approach introduces a novel Transformer-based decoder for interaction prediction, providing an explicit way to model the interactions between agents.

Regarding the utilization of prediction models for planning tasks, numerous works focus on multi-agent joint motion prediction frameworks \cite{sun2022intersim, gilles2022thomas, mo2022multi, jia2022multi} that enable efficient and consistent prediction of multi-modal multi-agent trajectories. An inherent issue in existing motion prediction models is that they often ignore the influence of the AV's actions, rendering them unsuitable for downstream planning tasks. To tackle this problem, several conditional multi-agent motion prediction models \cite{song2020pip, huang2022conditional, espinoza2022deep} have been proposed by integrating AV planning information into the prediction process. However, these models still exhibit one-way interactions, neglecting the mutual influence among agents. In contrast, our approach aims to jointly predict the future trajectories of surrounding agents and facilitate AV planning through iterative mutual interaction modeling. 

\subsection{Learning for Decision-making}
The primary objective of the motion prediction module is to enable the planning module to make safe and intelligent decisions. This can be achieved through the use of offline learning methods that can learn decision-making policies from large-scale driving datasets. Imitation learning stands as the most prevalent approach, which aims to learn a driving policy that can replicate expert behaviors \cite{huang2020multi, xu2022bits}. Offline reinforcement learning \cite{kumar2020conservative} has also gained interest as it combines the benefits of reinforcement learning and large collected datasets. However, direct policy learning lacks interpretability and safety assurance, and often suffers from distributional shifts. In contrast, planning with a learned motion prediction model is believed to be more interpretable and robust \cite{huang2022differentiable, casas2021mp3, zeng2019end, cui2021lookout}, making it a more desirable way for autonomous driving. Our proposed approach aims to enhance the capability of prediction models that can improve interactive decision-making performance.

\begin{figure*}
    \centering
    \includegraphics[width=0.86\linewidth]{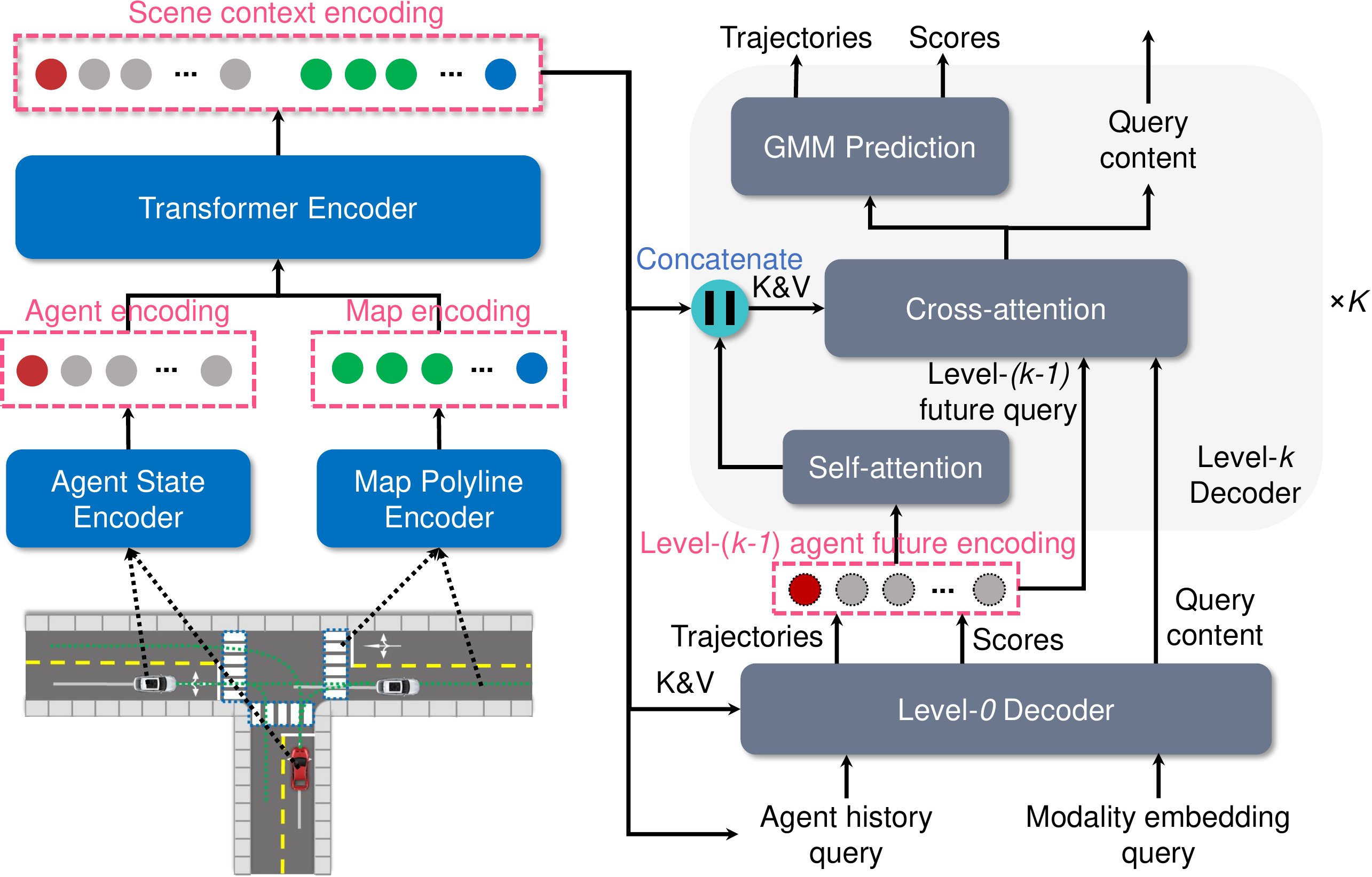}
    \caption{Overview of our proposed {\em GameFormer} framework. The scene context encoding is obtained via a Transformer-based encoder; the level-$0$ decoder takes the modality embedding and agent history encodings as query and outputs level-$0$ future trajectories and scores; the level-$k$ decoder uses a self-attention module to model the level-$(k-1$) future interaction and append it to the scene context encoding. }
    \vspace{-0.4cm}
    \label{fig:2}
\end{figure*}

\section{GameFormer}
\label{sec:method}
We introduce our interactive prediction and planning framework, called {\em GameFormer}, which adopts the Transformer encoder-decoder architecture (see Fig. \ref{fig:2}). In the following sections, we first define the problem and discuss the level-$k$ game theory that guides the design of the model and learning process in Sec. \ref{sec:game}. We then describe the encoder component of the model, which encodes the scene context, in Sec. \ref{sec:encode}, and the decoder component, which incorporates a novel interaction modeling concept, in Sec. \ref{sec:decode}. Finally, we present the learning process that accounts for interactions among different reasoning levels in Sec. \ref{sec:learning}.

\subsection{Game-theoretic Formulation}
\label{sec:game}
We consider a driving scene with $N$ agents, where the AV is denoted as $A_{0}$ and its neighboring agents as $A_{1}, \cdots, A_{N-1}$ at the current time $t=0$. Given the historical states of all agents (including the AV) over an observation horizon $T_h$, $\mathbf{S} = \{ \mathbf{s}_{i}^{-T_h:0} \}$, as well as the map information $\mathbf{M}$ including traffic lights and road waypoints, the goal is to jointly predict the future trajectories of neighboring agents $\mathbf{Y}^{1:T_f}_{1:N-1}$ over the future horizon $T_f$, as well as a planned trajectory for the AV $\mathbf{Y}^{1:T_f}_{0}$. In order to capture the uncertainty, the results are multi-modal future trajectories for the AV and neighboring agents, denoted by $\mathbf{Y}^{1:T_f}_{i} = \{ \mathbf{y}^{1:T_f}_{j}, \ p_j | j = \! 1:M \}$, where $\mathbf{y}^{1:T_f}_{j}$ is a sequence of predicted states, $p_j$ the probability of the trajectory, and $M$ the number of modalities.

We leverage level-$k$ game theory to model agent interactions in an iterative manner. Instead of simply predicting a single set of trajectories, we predict a hierarchy of trajectories to model the cognitive interaction process. At each reasoning level, with the exception of level-$0$, the decoder takes as input the prediction results from the previous level, which effectively makes them a part of the scene, and estimates the responses of agents in the current level to other agents in the previous level. We denote the predicted multi-modal trajectories (essentially a Gaussian mixture model) of agent $i$ at reasoning level $k$ as $\pi_{i}^{(k)}$, which can be regarded as a policy for that agent. The policy $\pi_{i}^{(k)}$ is conditioned on the policies of all other agents except the $i$-th agent at level-$(k-1)$, denoted by $\pi_{\neg i}^{(k-1)}$. For instance, the AV's policy at level-$2$ $\pi_0^{(2)}$ would take into account all neighboring agents' policies at level-$1$ $\pi_{1:N-1}^{(1)}$. Formally, the $i$-th agent’s level-$k$ policy is set to optimize the following objective:
\begin{equation}
\min_{\pi_i} \ \mathcal{L}^k_i \left( \pi_{i}^{(k)} \mid \pi_{\neg i}^{(k-1)} \right),    
\end{equation}
where $\mathcal{L}(\cdot)$ is the loss (or cost) function. It is important to note that policy $\pi$ here represents the multi-modal predicted trajectories (GMM) of an agent and that the loss function is calculated on the trajectory level.

For the level-$0$ policies, they do not take into account probable actions or reactions of other agents and instead behave independently. Based on the level-$k$ game theory framework, we design the future decoder, which we elaborate upon in Section \ref{sec:decode}.

\subsection{Scene Encoding}
\label{sec:encode}

\textbf{Input representation}. 
The input data comprises historical state information of agents, $S_p \in \mathbb{R}^{N \times T_h \times d_s}$, where $d_s$ represents the number of state attributes, and local vectorized map polylines $M \in \mathbb{R}^{N \times N_m \times N_p \times d_p}$. For each agent, we find $N_m$ nearby map elements such as routes and crosswalks, each containing $N_p$ waypoints with $d_p$ attributes. The inputs are normalized according to the state of the ego agent, and any missing positions in the tensors are padded with zeros.

\textbf{Agent History Encoding}. 
We use LSTM networks to encode the historical state sequence $S_p$ for each agent, resulting in a tensor $A_p \in \mathbb{R}^{N \times D}$, which contains the past features of all agents. Here, $D$ denotes the hidden feature dimension.

\textbf{Vectorized Map Encoding}. 
To encode the local map polylines of all agents, we use the multi-layer perceptron (MLP) network, which generates a map feature tensor $M_p \in \mathbb{R}^{N \times N_m \times N_p \times D}$ with a feature dimension of $D$. We then group the waypoints from the same map element and use max-pooling to aggregate their features, reducing the number of map tokens. The resulting map feature tensor is reshaped into $M_r \in \mathbb{R}^{N \times N_{mr} \times D}$, where $N_{mr}$ represents the number of aggregated map elements.

\textbf{Relation Encoding}. 
We concatenate the agent features and their corresponding local map features to create an agent-wise scene context tensor $C^{i} = [A_p, M_p^i] \in \mathbb{R}^{(N+N_{mr}) \times D}$ for each agent. We use a Transformer encoder with $E$ layers to capture the relationships among all the scene elements in each agent's context tensor $C^{i}$. The Transformer encoder is applied to all agents, generating a final scene context encoding $C_s \in \mathbb{R}^{N \times (N+N_{mr}) \times D}$, which represents the common environment background inputs for the subsequent decoder network.

\subsection{Future Decoding with Level-$k$ Reasoning}
\label{sec:decode}

\textbf{Modality embedding}. 
To account for future uncertainties, we need to initialize the modality embedding for each possible future, which serves as the query to the level-$0$ decoder. This can be achieved through either a heuristics-based method, learnable initial queries \cite{nayakanti2022wayformer}, or through a data-driven method \cite{shi2022motion}. Specifically, a learnable initial modality embedding tensor $I \in \mathbb{R}^{N \times M \times D}$ is generated, where $M$ represents the number of future modalities.

\textbf{Level-$0$ Decoding}. 
In the level-$0$ decoding layer, a multi-head cross-attention Transformer module is utilized, which takes as input the combination of the initial modality embedding $I$ and the agent's historical encoding in the final scene context $C_{s, A_p}$ (by inflating a modality axis), resulting in $(C_{s, A_p}+I) \in \mathbb{R}^{N \times M \times D}$ as the query and the scene context encoding $C_s$ as the key and value. The attention is applied to the modality axis for each agent, and the query content features can be obtained after the attention layer as $Z_{L_0} \in \mathbb{R}^{N \times M \times D}$. Two MLPs are appended to the query content features $Z_{L_0}$ to decode the GMM components of predicted futures $G_{L_0} \in \mathbb{R}^{N \times M \times T_f \times 4}$ (corresponding to ($\mu_x, \mu_y, \log \sigma_x, \log \sigma_y)$ at every timestep) and the scores of these components $P_{L_0} \in \mathbb{R}^{N \times M \times 1}$.

\begin{figure}[htp]
    \centering
    \includegraphics[width=\linewidth]{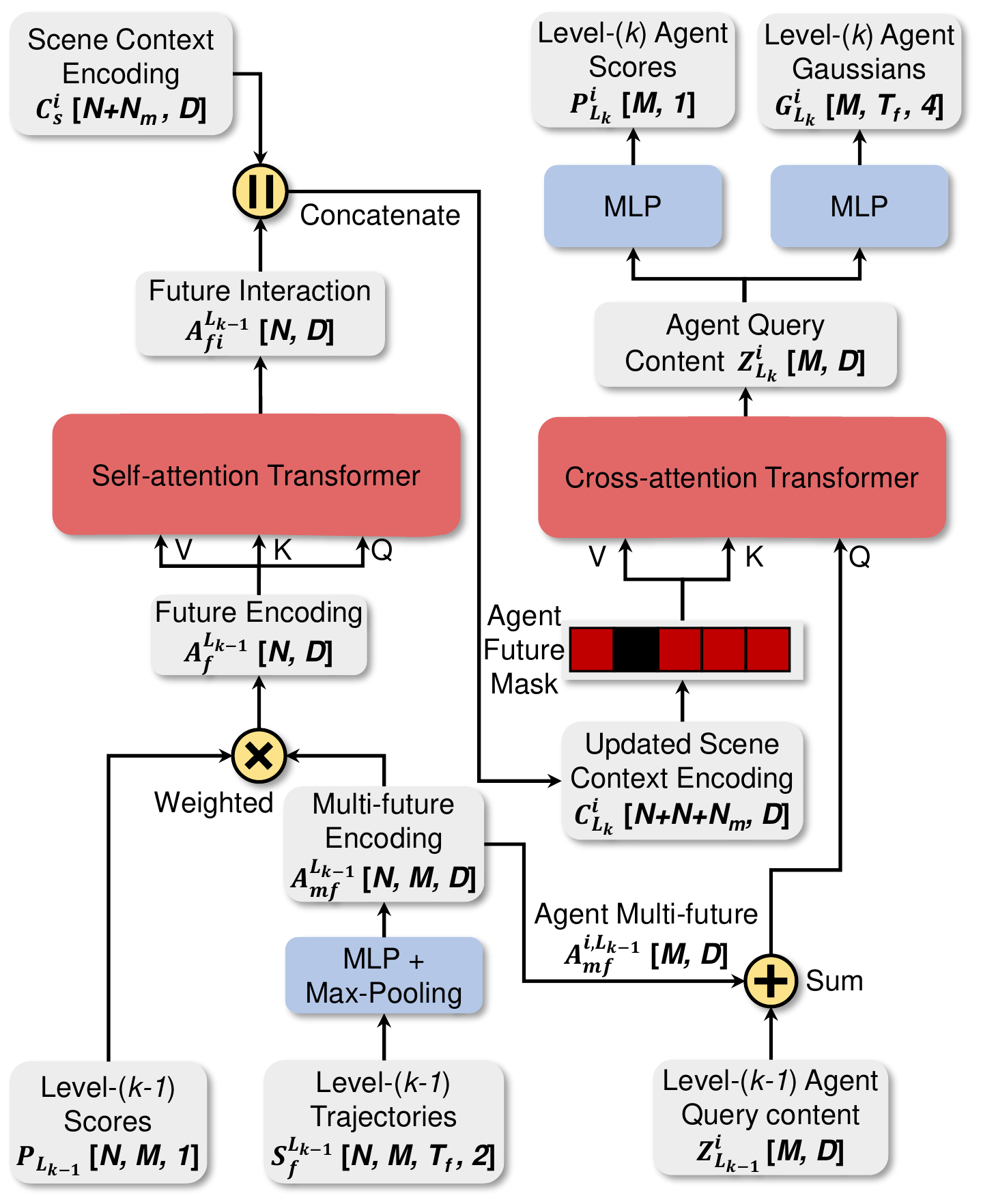}
    \caption{The detailed structure of a level-$k$ interaction decoder.}
    \vspace{-0.2cm}
    \label{fig:3}
\end{figure}

\textbf{Interaction Decoding}. 
The interaction decoding stage contains $K$ decoding layers corresponding to $K$ reasoning levels. In the level-$k$ layer ($k \ge 1$), it receives all agents' trajectories from the level-($k-1$) layer $S^{L_{k-1}}_f \in \mathbb{R}^{N \times M \times T_f \times 2}$ (the mean values of the GMM $G_{L_{k-1}}$) and use an MLP with max-pooling on the time axis to encode the trajectories, resulting in a tensor of agent multi-modal future trajectory encoding $A^{L_{k-1}}_{mf} \in \mathbb{R}^{N \times M \times D}$. Then, we apply weighted-average-pooling on the modality axis with the predicted scores from the level-($k-1$) layer $P_{L_{k-1}}$ to obtain the agent future features $A^{L_{k-1}}_{f} \in \mathbb{R}^{N \times D}$. We use a multi-head self-attention Transformer module to model the interactions between agent future trajectories $A_{fi}^{L_{k-1}}$ and concatenate the resulting interaction features with the scene context encoding from the encoder part. This yields an updated scene context encoding for agent $i$, denoted by $C^{i}_{L_{k}} = [A_{fi}^{L_{k-1}}, C_s^i] \in \mathbb{R}^{(N+N_m+N) \times D}$. We adopt a multi-head cross-attention Transformer module with the query content features from the level-($k-1$) layer $Z^i_{L_{k-1}}$ and agent future features $A^{L_{k-1}}_{mf}$, $(Z^i_{L_{k-1}}+A^{i, L_{k-1}}_{mf}) \in \mathbb{R}^{M \times D}$ as query and the updated scene context encoding $C^{i}_{L_{k}}$ as key and value. We use a masking strategy to prevent an agent from accessing its own future information from the last layer. For example, agent $A_0$ can only get access to the future interaction features of other agents $\{A_1, \cdots, A_{N-1}\}$. Finally, the resulting query content tensor from the cross-attention module $Z^i_{L_{k}}$ is passed through two MLPs to decode the agent's GMM components and scores, respectively. Fig. \ref{fig:3} illustrates the detailed structure of a level-$k$ interaction decoder. Note that we share the level-$k$ decoder for all agents to generate multi-agent trajectories at that level. At the final level of interaction decoding, we can obtain multi-modal trajectories for the AV and neighboring agents $G_{L_K}$, as well as their scores $P_{L_K}$.

\subsection{Learning Process}
\label{sec:learning}
We present a learning process to train our model using the level-$k$ game theory formalism. First, we employ imitation loss as the primary loss to regularize the agent's behaviors, which can be regarded as a surrogate for factors such as traffic regulations and driving styles. The future behavior of an agent is modeled as a Gaussian mixture model (GMM), where each mode $m$ at time step $t$ is described by a Gaussian distribution over the $(x, y)$ coordinates, characterized by mean $\mu_m^t$ and covariance $\sigma_m^t$. The imitation loss is computed using the negative log-likelihood loss from the best-predicted component $m^*$ (closest to the ground truth) at each timestep, as formulated:
\begin{equation}
\mathcal{L}_{IL} = \sum_{t=1}^{T_f} \mathcal{L}_{NLL}(\mu^{t}_{m^*}, \sigma^{t}_{m^*}, p_{m*}, \mathbf{s}_t).
\end{equation}

The negative log-likelihood loss function $\mathcal{L}_{NLL}$ is defined as follows:
\begin{equation}
\small
\mathcal{L}_{NLL} = \log \sigma_x + \log \sigma_y + \frac{1}{2} \left( \left( \frac{dx}{\sigma_x} \right)^2 + \left( \frac{dx}{\sigma_x} \right)^2 \right) - \log (p_{m*}),
\end{equation}
where $d_x = \mathbf{s}_x - \mu_x$ and $d_y = \mathbf{s}_y - \mu_y$, $(\mathbf{s}_x, \mathbf{s}_y)$ is ground-truth position; $p_{m*}$ is the probability of the selected component, and we use the cross-entropy loss in practice.

For a level-$k$ agent $A_i^{(k)}$, we design an auxiliary loss function inspired by prior works \cite{liu2021deep, hanselmann2022king, chen2022scept} that considers the agent's interactions with others. The safety of agent interactions is crucial, and we use an interaction loss (applicable only to decoding levels $k\ge1$) to encourage the agent to avoid collisions with the possible future trajectories of other level-$(k-1)$ agents. Specifically, we use a repulsive potential field in the interaction loss to discourage the agent's future trajectories from getting too close to any possible trajectory of any other level-$(k-1)$ agent $A_{\neg i}^{(k-1)}$. The interaction loss is defined as follows:
\begin{equation}
\mathcal{L}_{Inter} = \sum_{m=1}^{M} \sum_{t=1}^{T_f} \max_{ \substack{\forall j \neq i \\ \forall n \in {1:M}}} \frac{1}{ d \left( \mathbf{\hat s}^{(i, k)}_{m, t}, \mathbf{\hat s}^{(j, k-1)}_{n, t} \right) + 1},
\end{equation}
where $d(\cdot, \cdot)$ is the $L_2$ distance between the future states ($(x, y)$ positions), $m$ is the mode of the agent $i$, $n$ is the mode of the level-$(k-1)$ agent $j$. To ensure activation of the repulsive force solely within close proximity, a safety margin is introduced, meaning the loss is only applied to interaction pairs with distances smaller than a threshold.

The total loss function for the level-$k$ agent $i$ is the weighted sum of the imitation loss and interaction loss.
\begin{equation}
\label{loss}
\mathcal{L}_{i}^{k}(\pi_i^{(k)}) = w_{1} \mathcal{L}_{IL}(\pi_i^{(k)}) + w_{2} \mathcal{L}_{Inter}(\pi_i^{(k)}, \pi_{\neg i}^{(k-1)}),
\end{equation}
where $w_1$ and $w_2$ are the weighting factors to balance the influence of the two loss terms.

\section{Experiments}
\subsection{Experimental Setup}
\textbf{Dataset}. 
We set up two different model variants for different evaluation purposes. The prediction-oriented model is trained and evaluated using the Waymo open motion dataset (WOMD) \cite{ettinger2021large}, specifically addressing the task of predicting the joint trajectories of two interacting agents. For the planning tasks, we train and test the models on both WOMD with selected interactive scenarios and the nuPlan dataset \cite{nuplan} with a comprehensive evaluation benchmark.

\textbf{Prediction-oriented model}. 
We adopt the setting of the WOMD interaction prediction task, where the model predicts the joint future positions of two interacting agents 8 seconds into the future. The neighboring agents within the scene will serve as the background information in the encoding stage, while only the two labeled interacting agents' joint future trajectories are predicted. The model is trained on the entire WOMD training dataset, and we employ the official evaluation metrics, which include minimum average displacement error (minADE), minimum final displacement error (minFDE), miss rate, and mean average precision (mAP). We investigate two different prediction model settings. Firstly, we consider the joint prediction setting, where only $M=6$ joint trajectories of the two agents are predicted \cite{ngiam2021scene}. Secondly, we examine the marginal prediction setting and train our model to predict $M=64$ marginal trajectories for each agent in the interaction pair. During inference, the EM method proposed in MultiPath++ \cite{varadarajan2022multipath++} is employed to generate a set of $6$ marginal trajectories for each agent, from which the top $6$ joint predictions are selected for these two agents.

\textbf{Planning-oriented model}.
We introduce another model variant designed for planning tasks. Specifically, this variant takes into account multiple neighboring agents around the AV and predicts their future trajectories. The model is trained and tested across two datasets: WOMD and nuPlan. For WOMD, we randomly select 10,000 20-second scenarios, where 9,000 of them are used for training and the remaining 1,000 for validation. Then, we evaluate the model's joint prediction and planning performance on 400 9-second interactive and dynamic scenarios (\eg, lane-change, merge, and left-turn) in both open-loop and closed-loop settings. To conduct closed-loop testing, we utilize a log-replay simulator \cite{huang2022differentiable} to replay the original scenarios involving other agents, with our planner taking control of the AV. In open-loop testing, we employ distance-based error metrics, which include planning ADE, collision rate, miss rate, and prediction ADE. In closed-loop testing, we focus on evaluating the planner's performance in a realistic driving context by measuring metrics including success rate (no collision or off-route), progress along the route, longitudinal acceleration and jerk, lateral acceleration, and position errors. For the nuPlan dataset, we design a comprehensive planning framework and adhere to the nuPlan challenge settings to evaluate the planning performance. Specifically, we evaluate the planner's performance in three tasks: open-loop planning, closed-loop planning with non-reactive agents, and closed-loop with reactive agents. These tasks are evaluated using a comprehensive set of metrics provided by the nuPlan platform, and an overall score is derived based on these tasks. More information about our models is provided in the supplementary material.

\subsection{Main Results}
\subsubsection{Interaction Prediction}
Within the prediction-oriented model, we use a stack of $E=6$ Transformer encoder layers, and the hidden feature dimension is set to $D=256$. We consider $20$ neighboring agents around the two interacting agents as background information and employ $K=6$ decoding layers. The model only generates trajectories for the two labeled interacting agents.  Moreover, the local map elements for each agent comprise possible lane polylines and crosswalk polylines.

\textbf{Quantitative results}. 
Table \ref{tab1} summarizes the prediction performance of our model in comparison with state-of-the-art methods on the WOMD interaction prediction (joint prediction of two interacting agents) benchmark. The metrics are averaged over different object types (vehicle, pedestrian, and cyclist) and evaluation times (3, 5, and 8 seconds). Our joint prediction model (GameFormer (J, $M$=6)) outperforms existing methods in terms of position errors. This can be attributed to its superior ability to capture future interactions between agents through an iterative process and to predict future trajectories in a scene-consistent manner. However, the scoring performance of the joint model is limited without predicting an over-complete set of trajectories and aggregation. To mitigate this issue, we employ the marginal prediction model (GameFormer (M, $M$=64)) with EM aggregation, which significantly improves the scoring performance (better mAP metric). The overall performance of our marginal model is comparable to that of the ensemble and more complicated MTR model \cite{shi2022motion}. Nevertheless, it is worth noting that marginal ensemble models may not be practical for real-world applications due to their substantial computational burden. Therefore, we utilize the joint prediction model, which provides better prediction accuracy and computational efficiency, for planning tests.

\begin{table}[htp]
\caption{Comparison with state-of-the-art models on the WOMD interaction prediction benchmark}
\centering
\resizebox{\linewidth}{!}{%
\begin{tabular}{@{}lcccc@{}}
\toprule
Model                                   & minADE ($\downarrow$) & minFDE ($\downarrow$) & Miss rate ($\downarrow$) & mAP ($\uparrow$) \\ \midrule
LSTM baseline \cite{ettinger2021large}  & 1.9056                &  5.0278               &  0.7750                  & 0.0524    \\
Heat \cite{mo2022multi}                 & 1.4197                &  3.2595               &  0.7224                  & 0.0844    \\
AIR$^2$ \cite{wu2021air}                & 1.3165                &  2.7138               &  0.6230                  & 0.0963    \\
SceneTrans \cite{ngiam2021scene}        & 0.9774                &  2.1892               &  0.4942                  & 0.1192     \\
DenseTNT \cite{gu2021densetnt}          & 1.1417                &  2.4904               &  0.5350                  & 0.1647   \\
M2I \cite{sun2022m2i}                   & 1.3506                &  2.8325               &  0.5538                  & 0.1239     \\
MTR \cite{shi2022motion}                & 0.9181                & 2.0633                &  \textbf{0.4411}         & \textbf{0.2037} \\ \midrule
GameFormer (M, $M$=64)                  & 0.9721                & 2.2146                &  0.4933                  & 0.1923 \\
GameFormer (J, $M$=6)                   & \textbf{0.9161}       & \textbf{1.9373}       &  0.4531                  & 0.1376 \\ \bottomrule
\end{tabular}
}
\vspace{-0.2cm}
\label{tab1}
\end{table}

\begin{figure*}[htp]
    \centering
    \includegraphics[width=0.93\linewidth]{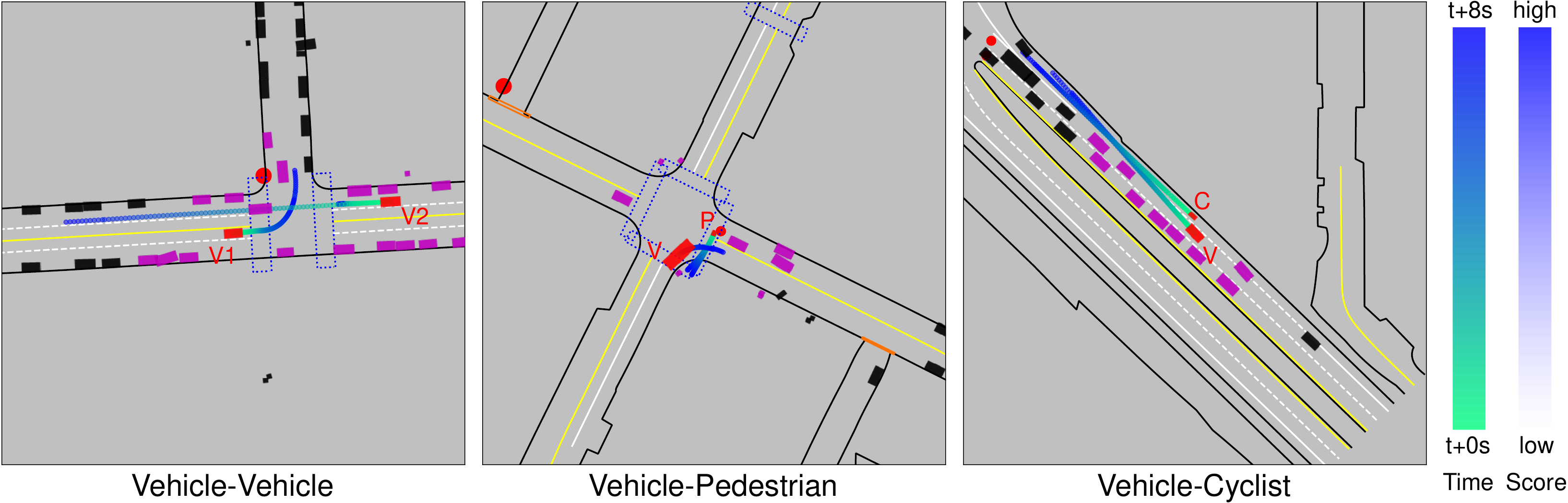}
    \caption{Qualitative results of the proposed method in interaction prediction (multi-modal joint prediction of two interacting agents). The red boxes are interacting agents to predict and the magenta boxes are background neighboring agents.}
    \label{fig:4}
\end{figure*}

\textbf{Qualitative results}. 
Fig. \ref{fig:4} illustrates the interaction prediction performance of our approach in several typical scenarios. In the vehicle-vehicle interaction scenario, two distinct situations are captured by our model: vehicle 2 accelerates to take precedence at the intersection, and vehicle 2 yields to vehicle 1. In both cases, our model predicts that vehicle 1 creeps forward to observe the actions of vehicle 2 before executing a left turn. In the vehicle-pedestrian scenario, our model predicts that the vehicle will stop and wait for the pedestrian to pass before starting to move. In the vehicle-cyclist interaction scenario, where the vehicle intends to merge into the right lane, our model predicts the vehicle will decelerate and follow behind the cyclist in that lane. Overall, the results manifest that our model can capture multiple interaction patterns of interacting agents and accurately predict their possible joint futures.

\begin{figure*}[htp]
    \centering
    \includegraphics[width=0.93\linewidth]{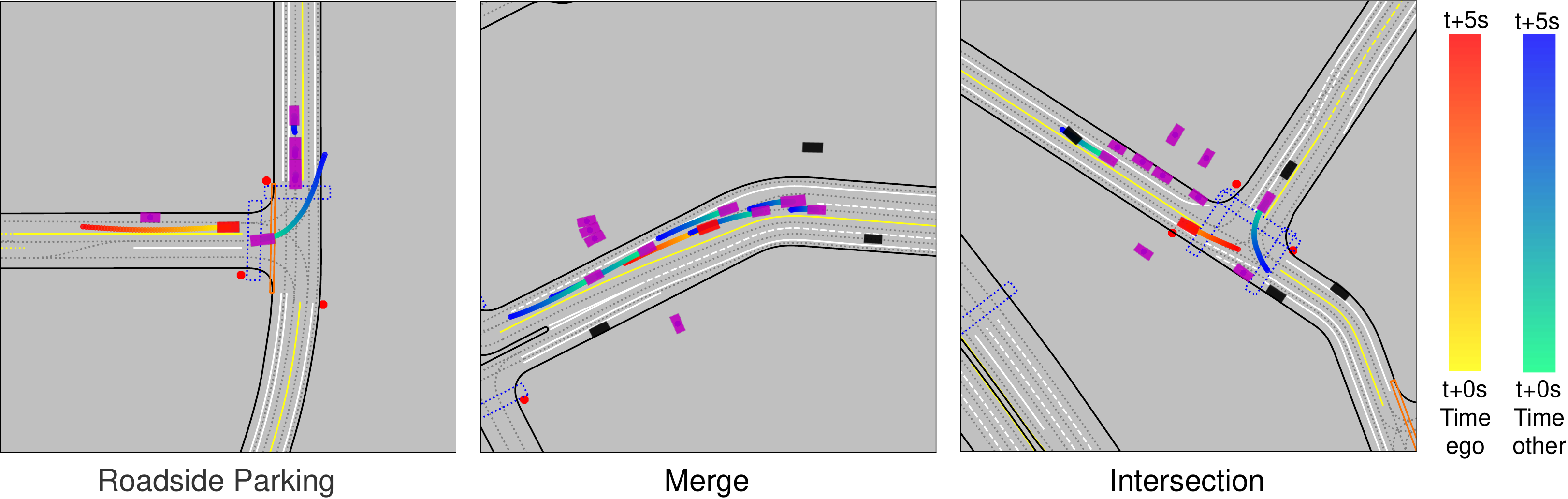}
    \caption{Qualitative results of the proposed method in open-loop planning. The red box is the AV and the magenta boxes are its neighboring agents; the red trajectory is the plan of the AV and the blue ones are the predictions of neighboring agents.}
    \label{fig:5}
    \vspace{-0.4cm}
\end{figure*}

\subsubsection{Open-loop Planning}
We first conduct the planning tests in selected WOMD scenarios with a prediction/planning horizon of 5 seconds. The model uses a stack of $E=6$ Transformer encoder layers, and we consider $10$ neighboring agents closest to the ego vehicle to predict $M=6$ joint future trajectories for them. 

\textbf{Determining the decoding levels}. 
To determine the optimal reasoning levels for planning, we analyze the impact of decoding layers on open-loop planning performance, and the results are presented in Table \ref{tab:2.5}. Although the planning ADE and prediction ADE exhibit a slight decrease with additional decoding layers, the miss rate and collision rate are at their lowest when the decoding level is $4$. The intuition behind this observation is that humans are capable of performing only a limited depth of reasoning, and the optimal iteration depth empirically appears to be $4$ in this test.

\begin{table}[htp]
\caption{Influence of decoding levels on open-loop planning}
\centering
\resizebox{\linewidth}{!}{
\begin{tabular}{@{}lcccc@{}} 
\toprule 		
Level       & Planning ADE      & Collision Rate    & Miss Rate         & Prediction ADE \\ \midrule
0           & 0.9458            & 0.0384            & 0.1154            & 1.0955  \\
1           & 0.8846            & 0.0305            & 0.0994            & 0.9377  \\
2           & 0.8529            & 0.0277            & 0.0897            & 0.8875  \\
3           & 0.8423            & 0.0269            & 0.0816            & 0.8723  \\
4           & 0.8329            & \textbf{0.0198}   & \textbf{0.0753}   & 0.8527    \\
5           & \textbf{0.8171}   & 0.0245            & 0.0777            & 0.8361    \\
6           & 0.8208            & 0.0238            & 0.0826            & \textbf{0.8355}    \\ \bottomrule
\end{tabular}
}
\label{tab:2.5}
\vspace{-0.3cm}
\end{table}

\textbf{Quantitative results}. 
Our joint prediction and planning model employs $4$ decoding layers, and the results of the final decoding layer (the most-likely future evaluated by the trained scorer) are utilized as the plan for the AV and predictions for other agents. We set up some imitation learning-based planning methods as baselines, which are: 1) vanilla imitation learning (IL), 2) deep imitative model (DIM) \cite{rhinehart2019deep}, 3) MultiPath++ \cite{varadarajan2022multipath++} (which predicts multi-modal trajectories for the ego agent), 4) MTR-e2e (end-to-end variant with learnable motion queries) \cite{shi2022motion}, and 5) differentiable integrated prediction and planning (DIPP) \cite{huang2022differentiable}. Table \ref{tab2} reports the open-loop planning performance of our model in comparison with the baseline methods. The results reveal that our model performs significantly better than vanilla IL and DIM, because they are just trained to output the ego's trajectory while not explicitly predicting other agents' future behaviors. Compared to performant motion prediction models (MultiPath++ and MTR-e2e), our model also shows better planning metrics for the ego agent. Moreover, our model outperforms DIPP (a joint prediction and planning method) in both planning and prediction metrics, especially the collision rate. These results emphasize the advantage of our model, which explicitly considers all agents' future behaviors and iteratively refines the interaction process. 

\begin{table*}[htp]
\caption{Evaluation of open-loop planning performance in selected WOMD scenarios}
\centering
\small
\begin{tabular}{@{}l|c|c|ccc|cc@{}}
\toprule
\multirow{2}{*}{Method} &
  \multirow{2}{*}{Collision rate (\%)} &
  \multirow{2}{*}{Miss rate (\%)} &
  \multicolumn{3}{c|}{Planning error (m)} &
  \multicolumn{2}{c}{Prediction error (m)} \\
                &               &                 & @1s            & @3s            & @5s           & ADE           & FDE \\ \midrule
Vanilla IL     & 4.25          & 15.61            & 0.216          & 1.273          & 3.175          & --            & --       \\
DIM            & 4.96          & 17.68            & 0.483          & 1.869          & 3.683          & --            & --         \\
MultiPath++    & 2.86          & 8.61             & 0.146          & 0.948          & 2.719          & --            & --       \\
MTR-e2e        & 2.32          & 8.88             & 0.141          & 0.888          & 2.698          & --            & --     \\
DIPP           & 2.33          & 8.44             & 0.135          & 0.928          & 2.803          & 0.925         & 2.059       \\
Ours           & \textbf{1.98} & \textbf{7.53}    & \textbf{0.129} &\textbf{0.836}  & \textbf{2.451} &\textbf{0.853} & \textbf{1.919} \\ \bottomrule
\end{tabular}
\label{tab2}
\end{table*}

\begin{table*}[htp]
\caption{Evaluation of closed-loop planning performance in selected WOMD scenarios}
\centering
\resizebox{\linewidth}{!}{
\begin{tabular}{@{}l|cc|ccc|ccc@{}}
\toprule
\multirow{2}{*}{Method} &
\multicolumn{1}{c}{Success rate} &
\multicolumn{1}{c|}{Progress} &
\multicolumn{1}{c}{Acceleration} &
\multicolumn{1}{c}{Jerk} &
\multicolumn{1}{c|}{Lateral acc.} &
\multicolumn{3}{c}{Position error to expert driver ($m$)} \\
                    &  (\%)                 & $(m)$                 & ($m/s^2$)             & ($m/s^3$)             & ($m/s^2$)             & @3s                   & @5s            & @8s \\ \midrule
Vanilla IL          & 0                     & 6.23                  & 1.588                 & 16.24                 & 0.661                 & 9.355                 & 20.52          & 46.33  \\
RIP                 & 19.5                  & 12.85                 & 1.445                 & 14.97                 & 0.355                 & 7.035                 & 17.13          & 38.25  \\
CQL                 & 10                    & 8.28                  & 3.158                 & 25.31                 & 0.152                 & 10.86                 & 21.18          & 40.17   \\ \midrule
DIPP                & 68.12$\pm$5.51        & 41.08$\pm$5.88        & 1.44$\pm$0.18         & 12.58$\pm$3.23        & 0.31$\pm$0.11         & 6.22$\pm$0.52         & 15.55$\pm$1.12 & 26.10$\pm$3.88 \\
Ours                & 73.16$\pm$6.14        & 44.94$\pm$7.69        & 1.19$\pm$0.15         & 13.63$\pm$2.88        & 0.32$\pm$0.09         & 5.89$\pm$0.78         & 12.43$\pm$0.51 & 21.02$\pm$2.48 \\
DIPP (w/ refinement)& 92.16$\pm$0.62        &51.85$\pm$0.14         & 0.58$\pm$0.03         & \textbf{1.54}$\pm$0.19& 0.11$\pm$0.01         & 2.26$\pm$0.10         & 5.55$\pm$0.24  & 12.53$\pm$0.48 \\
Ours (w/ refinement)&\textbf{94.50}$\pm$0.66&\textbf{52.67}$\pm$0.33& \textbf{0.53}$\pm$0.02& 1.56$\pm$0.23         & \textbf{0.10}$\pm$0.01&\textbf{2.11}$\pm$0.21 & \textbf{4.87}$\pm$0.18 &\textbf{11.13}$\pm$0.33  \\ \bottomrule
\end{tabular}
}
\vspace{-0.3cm}
\label{tab3}
\end{table*}

\textbf{Qualitative results}.
Fig. \ref{fig:5} displays qualitative results of our model's open-loop planning performance in complex driving scenarios. For clarity, only the most-likely trajectories of the agents are displayed. These results demonstrate that our model can generate a plausible future trajectory for the AV and handle diverse interaction scenarios, and predictions of the surrounding agents enhance the interpretability of our planning model's output.

\subsubsection{Closed-loop Planning}
We evaluate the closed-loop planning performance of our model in selected WOMD scenarios. Within a simulated environment \cite{huang2022differentiable}, we execute the planned trajectory generated by the model and update the ego agent's state at each time step, while other agents follow their logged trajectories from the dataset. Since other agents do not react to the ego agent, the success rate is a lower bound for safety assessment. For planning-based methods (DIPP and our proposed method), we project the output trajectory onto a reference path to ensure the ego vehicle's adherence to the roadway. Additionally, we employ a cost-based refinement planner \cite{huang2022differentiable}, which utilizes the initial output trajectory and the predicted trajectories of other agents to explicitly regulate the ego agent's actions. Our method is compared against four baseline methods: 1) vanilla IL, 2) robust imitative planning (RIP) \cite{filos2020can}, 3) conservative Q-learning (CQL) \cite{kumar2020conservative}, and 4) DIPP \cite{huang2022differentiable}. We report the means and standard deviations of the planning-based methods over three training runs (models trained with different seeds). The quantitative results of closed-loop testing are summarized in Table \ref{tab3}. The results show that the IL and offline RL methods exhibit subpar performance in the closed-loop test, primarily due to distributional shifts and casual confusion. In contrast, planning-based methods perform significantly better across all metrics. Without the refinement step, our model outperforms DIPP because it captures agent interactions more effectively and thus the raw trajectory is closer to an expert driver. With the refinement step, the planner becomes more robust against training seeds, and our method surpasses DIPP because it can deliver better predictions of agent interactions and provide a good initial plan to the refinement planner.

\subsubsection{nuPlan Benchmark Evaluation} 
To handle diverse driving scenarios in the nuPlan platform \cite{nuplan}, we develop a comprehensive planning framework {\em GameFormer Planner}. It fulfills all important steps in the planning pipeline, including feature processing, path planning, model query, and motion refinement. We increase the prediction and planning horizon to 8 seconds to meet benchmark requirements. The evaluation is conducted over three tasks: open-loop (OL) planning, closed-loop (CL) planning with non-reactive agents, and closed-loop planning with reactive agents. The score for each individual task is calculated using various metrics and scoring functions, and an overall score is obtained by aggregating these task-specific scores. It is important to note that we reduce the size of our model (encoder and decoder layers) due to limited computational resources on the test server. The performance of our model on the nuPlan test benchmark is presented in Table \ref{tab5}, in comparison with other competitive learning-based methods and a rule-based approach (IDM Planner). The results reveal the capability of our planning framework in achieving high-quality planning results across the evaluated tasks. Moreover, the closed-loop visualization results illustrate the ability of our model to facilitate the ego vehicle in making interactive and human-like decisions. 

\begin{table}[htp]
\caption{Results on the nuPlan planning test benchmark}
\centering
\resizebox{\linewidth}{!}{
\begin{tabular}{@{}lcccc@{}}
\toprule
Method            & Overall     & OL        & CL non-reactive   & CL reactive    \\ \midrule
Hoplan            & 0.8745      & 0.8523    & 0.8899            & 0.8813                   \\
Multi\_path       & 0.8477      & 0.8758    & 0.8165            & 0.8506                     \\
GameFormer        & 0.8288      & 0.8400    & 0.8087            & 0.8376                    \\
Urban Driver      & 0.7467      & 0.8629    & 0.6821            & 0.6952                    \\
IDM Planner       & 0.5912      & 0.2944    & 0.7243            & 0.7549                  \\ \bottomrule
\end{tabular}%
}
\label{tab5}
\vspace{-0.2cm}
\end{table}

\subsection{Ablation Study}
\textbf{Effects of agent future modeling}.
We investigate the impact of different agent future modeling settings on open-loop planning performance in WOMD scenarios. We compare our base model to three ablated models: 1) \textit{No future}: agent future trajectories from the preceding level are not incorporated in the decoding process at the current level, 2) \textit{No self-attention}: agent future trajectories are incorporated but not processed through a self-attention module, and 3) \textit{No interaction loss}: the model is trained without the proposed interaction loss. The results, as presented in Table \ref{tabc}, demonstrate that our game-theoretic approach can significantly improve planning and prediction accuracy. It underscores the advantage of utilizing the future trajectories of agents from the previous level as contextual information for the current level. Additionally, incorporating a self-attention module to represent future interactions among agents improves the accuracy of planning the prediction. Using the proposed interaction loss during training can significantly reduce the collision rate.

\begin{table}[htp]
\caption{Influence of future modeling on open-loop planning}
\resizebox{\linewidth}{!}{%
\begin{tabular}{@{}lcccc@{}}
\toprule
                    & Planning ADE      & Collision Rate    & Miss Rate         & Prediction ADE  \\ \midrule
No future           & 0.9210            &  0.0295           & 0.0963            &  0.9235 \\
No self-attention   & 0.8666            &  0.0231           & 0.0860            &  0.8856\\
No interaction loss & 0.8415            &  0.0417           & 0.0846            & \textbf{0.8486} \\
Base                & \textbf{0.8329}   & \textbf{0.0198}   & \textbf{0.0753}   &  0.8527   \\ \bottomrule
\end{tabular}%
}
\label{tabc}
\vspace{-0.1cm}
\end{table}

\textbf{Influence of decoder structures}.
We investigate the influence of decoder structures on the open-loop planning task in WOMD scenarios. Specifically, we examine two ablated models. First, we assess the importance of incorporating $k$ independent decoder layers, as opposed to training a single shared interaction decoder and iteratively applying it $k$ times. Second, we explore the impact of simplifying the decoder into a multi-layer Transformer that does not generate intermediate states. This translates into applying the loss solely to the final decoding layer, rather than all intermediate layers. The results presented in Table \ref{tabd} demonstrate better open-loop planning performance for the base model (independent decoding layers with intermediate trajectories). This design allows each layer to capture different levels of relationships, thereby facilitating hierarchical modeling. In addition, the omission of intermediate trajectory outputs can degrade the model's performance, highlighting the necessity of regularizing the intermediate state outputs.

\begin{table}[htp]
\centering
\caption{Influence of decoder structures on open-loop planning}
\resizebox{\linewidth}{!}{
\begin{tabular}{@{}lcccc@{}}
\toprule 		
                            & Planning ADE      & Collision Rate    & Miss Rate         & Prediction ADE \\ \midrule
Base                        & \textbf{0.8329}   & \textbf{0.0198}   & \textbf{0.0753}   & \textbf{0.8547}    \\
Shared decoder              & 0.9196            & 0.0382            & 0.0860            & 0.9095      \\
Multi-layer decoder         & 0.9584            & 0.0353            & 0.0988            & 0.9637      \\ \bottomrule
\end{tabular}
}
\label{tabd}
\vspace{-0.1cm}
\end{table}

\textbf{Ablation results on the interaction prediction task}.
We investigate the influence of the decoder on the WOMD interaction prediction task. Specifically, we vary the decoding levels from 0 to 8 to determine the optimal decoding level for this task. Moreover, we remove either the agent future encoding part from the decoder or the self-attention module (for modeling agent future interactions) to investigate their influences on prediction performance. We train the ablated models using the same training set and evaluate their performance on the validation set. The results in Table \ref{tabe} reveal that the empirically optimal number of decoding layers is 6 for the interaction prediction task. It is evident that fewer decoding layers fail to adequately capture the interaction dynamics, resulting in subpar prediction performance. However, using more than 6 decoding layers may introduce training instability and overfitting issues, leading to worse testing performance. Similarly, we find that incorporating predicted agent future information is crucial for achieving good performance, and using self-attention to model the interaction among agents' futures can also improve prediction accuracy. 

\begin{table}[htp]
\centering
\caption{Decoder ablation results on interaction prediction}
\resizebox{\linewidth}{!}{
\begin{tabular}{@{}lcccc@{}}
\toprule
Decoding layers     & minADE            & minFDE            & Miss Rate         & mAP    \\ \midrule
$K$=0               & 1.0505            & 2.2905            & 0.5113            & 0.1226 \\
$K$=1               & 1.0169            & 2.1876            & 0.5061            & 0.1281 \\
$K$=3               & 0.9945            & 2.1143            & 0.5026            & 0.1265 \\
$K$=6               & \textbf{0.9133}   & \textbf{1.9251}   & \textbf{0.4564}   & \textbf{0.1339} \\
$K$=8               & 0.9839            & 2.1515            & 0.5003            & 0.1255 \\ \midrule
$K$=6 w/o future    & 0.9862            & 2.0848            & 0.4979            & 0.1256 \\
$K$=6 w/o self-attention & 0.9263            & 1.9931            & 0.4599            & 0.1281 \\ \bottomrule
\end{tabular}
}
\label{tabe}
\vspace{-0.3cm}
\end{table}

\section{Conclusions}
This paper introduces {\em GameFormer}, a Transformer-based model that utilizes hierarchical game theory for interactive prediction and planning. Our proposed approach incorporates novel level-$k$ interaction decoders in the Transformer prediction model that iteratively refine the future trajectories of interacting agents. We also implement a learning process that regulates the predicted behaviors of agents based on the prediction results from the previous level. Experimental results on the Waymo open motion dataset demonstrate that our model achieves state-of-the-art accuracy in interaction prediction and outperforms baseline methods in both open-loop and closed-loop planning tests. Moreover, our proposed planning framework delivers leading performance on the nuPlan planning benchmark.

\section*{Acknowledgement}
This work was supported in part by the A*STAR AME Young Individual Research Grant (No. A2084c0156), the MTC Individual Research Grants (No.M22K2c0079), the ANR-NRF joint grant (No.NRF2021-NRF-ANR003 HM Science), and the SUG-NAP Grant of Nanyang Technological University, Singapore.

{\small
\bibliographystyle{ieee_fullname}
\bibliography{egbib}
}

\clearpage

\newcommand{\mathbbm}[1]{\text{\usefont{U}{bbm}{m}{n}#1}} 
\newcommand{\appendixhead}%
{\centering\textbf{\Large GameFormer: Game-theoretic Modeling and Learning of Transformer-based Interactive Prediction and Planning for Autonomous Driving \newline \newline Supplementary Material}
\vspace{0.3in}}

\counterwithin{figure}{section}
\counterwithin{table}{section}
\counterwithin{equation}{section}

\renewcommand\thefigure{S\arabic{figure}}   
\renewcommand\thetable{S\arabic{table}}   
\renewcommand\theequation{S\arabic{equation}}

\twocolumn[\appendixhead]

\appendix

\section{Experiment Details}
\subsection{Prediction-oriented Model}
\textbf{Model inputs}. 
In each scene, one of the two interacting agents is designated as the focal agent, with its current state serving as the origin of the coordinate system. We consider 10 surrounding agents closest to a target agent as the background agents, and therefore, there are two target agents to predict and up to 20 different background agents in a scene. The current and historical states of each agent are retrieved for the last one second at a sampling rate of 10Hz, resulting in a tensor with a shape of $(22\times11)$ for each agent. The state at each timestep includes the agent's position ($x,y$), heading angle ($\theta$), velocity ($v_x, v_y$), bounding box size ($L, W, H$), and one-hot category encoding of the agent (totally three types). All historical states for each agent are aggregated into a fixed-shape tensor of $(22\times11\times11)$, with missing agent states padded as zeros, to form the input tensor of historical agent states.

For each target agent, up to 6 drivable lanes (each extending 100 meters) that the agent may take are identified using depth-first search on the road graph, along with 4 nearby crosswalks as the local map context, with each map vector containing 100 waypoints. The features of a waypoint in a drivable lane include the position and heading angles of the centerline, left boundary, and right boundary, speed limit, as well as discrete attributes such as the lane type, traffic light state, and controlled by a stop sign. The features of a waypoint in the crosswalk polyline only encompass position and heading angle. Therefore, the local map context for a target agent comprises two tensors: drivable lanes with shape $(6\times100\times15)$ and crosswalks with shape $(4\times100\times3)$.

\textbf{Encoder structure}.
In the encoder part, we utilize two separate LSTMs to encode the historical states of the target and background agents, respectively, resulting in a tensor with shape $(22\times256)$ that encompasses all agents' historical state sequences. The local map context encoder consists of a lane encoder for processing the drivable lanes and a crosswalk encoder for the crosswalk polylines. The lane encoder employs MLPs to encode numeric features and embedding layers to encode discrete features, outputting a tensor of encoded lane vectors with shape $(2\times6\times100\times256)$, while the crosswalk encoder uses an MLP to encode numeric features, resulting in a tensor of crosswalk vectors with shape $(2\times4\times100\times256)$. Subsequently, we utilize a max-pooling layer (with a step size of 10) to aggregate the waypoints from a drivable lane in the encoded lane tensor, yielding a tensor with shape $(2\times6\times10\times256)$ that is reshaped to $(2\times60\times256)$. Similarly, the encoded crosswalk tensor is processed using a max-pooling layer with a step size of 20 to obtain a tensor with shape $(2\times20\times256)$. These two tensors are concatenated to produce an encoded local map context tensor with shape $(2\times80\times256)$. For each target agent, we concatenate its local map context tensor with the historical state tensor of all agents to obtain a scene context tensor with dimensions of $(102\times256)$, and we use self-attention Transformer encoder layers to extract the relationships among the elements in the scene. It is important to note that invalid positions in the scene context tensor are masked from attention calculations.

\textbf{Decoder structure}.
For the $M=6$ joint prediction model, we employ the learnable latent modality embedding with a shape of ($2\times6\times256$). For each agent, the query ($6\times256$) in the level-0 decoder is obtained by summing up the encoding of the target agent's history and its corresponding latent modality embedding; the value and key are derived from the scene context by the encoder. The level-0 decoder generates the multi-modal future trajectories of the target agent with $x$ and $y$ coordinates using an MLP from the attention output. The scores of each trajectory are decoded by another MLP with a shape of $(6\times1)$. In a level-$k$ decoder, we use a shared future encoder across different layers, which includes an MLP and a max-pooling layer, to encode the future trajectories from the previous level into a tensor with a shape of $(6\times256)$. Next, we employ the trajectory scores to average-pool the encoded trajectories, which results in the encoded future of the agent. The encoded futures of the two target agents are then fed into a self-attention Transformer layer to model their future interaction. Finally, the output of the Transformer layer is appended to the scene context obtained from the encoder.

For the $M=64$ marginal prediction model, we use a set of 64 fixed intention points that are encoded with MLPs to create the modality embedding with shape ($2\times64\times256$). This modality embedding serves as the query input for the level-0 decoder. The fixed intention points are obtained through the K-means method from the training dataset. For each scene, the intention points for the two target agents are normalized based on the focal agent's coordinate system. The other components of the decoder are identical to those used in the joint prediction model.

\textbf{Training}.
In the training dataset, each scene contains several agent tracks to predict, and we consider each track sequentially as the focal agent, while the closest track to the focal agent is chosen as the interacting agent. The task is to predict six possible joint future trajectories of these two agents. We employ only imitation loss at each level to improve the prediction accuracy and training efficiency. 

In the joint prediction model, we aim to predict the joint and scene-level future trajectories of the two agents. Therefore, we backpropagate the loss through the joint future trajectories of the two agents that most closely match the ground truth (i.e., have the least sum of displacement errors). In the marginal prediction model, we backpropagate the imitation loss to the individual agent through the positive GMM component that corresponds to the closest intention point to the endpoint of the ground-truth trajectory.

Our models are trained for 30 epochs using the AdamW optimizer with a weight decay of 0.01. The learning rate starts with 1e-4 and decays by a factor of 0.5 every 3 epochs after 15 epochs. We also clip the gradient norm of the network parameters with the max norm of the gradients as 5. We train the models using 4 NVIDIA Tesla V100 GPUs, with a batch size of 64 per GPU.

\textbf{Testing}. 
The testing dataset has three types of agents: vehicle, pedestrian, and cyclist. For the vehicle-vehicle interaction, we randomly select one of the two vehicles as the focal agent. For other types of interaction pairs (\eg, cyclist-vehicle and pedestrian-vehicle), we consider the cyclist or pedestrian as the focal agent. For the marginal prediction model, we employ the Expectation-Maximization (EM) method to aggregate trajectories for each agent. Specifically, we use the EM method to obtain 6 marginal trajectories (along with their probabilities) from the 64 trajectories predicted for each agent. Then, we consider the top 6 joint predictions from the 36 possible combinations of the two agents, where the confidence of each combination is the product of the marginal probabilities.

\subsection{Planning-oriented Model}
\textbf{Model inputs}.
In each scene, we consider the AV and 10 surrounding agents to perform planning for the AV and prediction for other agents. The AV's current state is the origin of the local coordinate system. The historical states of all agents in the past two seconds are extracted; for each agent, we find its nearby 6 drivable lanes and 4 crosswalks. Additionally, we extract the AV's traversed lane waypoints from its ground truth future trajectory and use a cubic spline to interpolate these waypoints to generate the AV's reference route. The reference route extends 100 meters ahead of the AV and contains 1000 waypoints with 0.1 meters intervals. It is represented as a tensor with shape $(1000\times5)$. The reference route tensor also contains information on the speed limit and stop points in addition to positions and headings.

\textbf{Model structure}.
For each agent, its scene context tensor is created as a concatenation of all agents' historical states and encoded local map elements, resulting in a tensor of shape $(91\times256)$. In the decoding stage, a learnable modality embedding of size $(11\times6\times256)$ and the agent's historical encoding are used as input to the level-0 decoder, which outputs six possible trajectories along with corresponding scores. In the level-$k$ decoder, the future encodings of all agents are obtained through a self-attention module of size $(11\times256)$, and are concatenated with the scene context tensor from the encoder. This concatenation generates an updated scene context tensor with a shape of $(102\times256)$. When decoding an agent's future trajectory at the current level, the future encoding of that agent in the scene context tensor is masked to avoid using its previously predicted future information. 

\textbf{Training}.
In data processing, we filter those scenes where the AV's moving distance is less than 5 meters (\eg, when stopping at a red light). Similarly, we perform joint future prediction and calculate the imitation loss through the joint future that is closest to the ground truth. The weights for the imitation loss and interaction loss are set to $w_1 = 1, w_2 = 0.1$. Our model is trained for 20 epochs using the AdamW optimizer with a weight decay of 0.01. The learning rate is initialized to 1e-4 and decreases by a factor of 0.5 every 2 epochs after the 10th epoch. We train the model using an NVIDIA RTX 3080 GPU, with a batch size of 32.

\textbf{Testing}.
The testing scenarios are extracted from the WOMD, wherein the ego agent shows dynamic driving behaviors\footnote{\url{https://github.com/smarts-project/smarts-project.offline-datasets/blob/master/waymo_candid_list.csv}}. In open-loop testing, we check collisions between the AV's planned trajectory and other agents' ground-truth future trajectories, and we count a miss if the distance between AV's planned state at the final step and the ground-truth state is larger than 4.5 meters. The planning errors and prediction errors are calculated according to the most-likely trajectories scored by the model. In closed-loop testing, the AV plans a trajectory at every timestep with an interval of 0.1 seconds and executes the first step of the plan.

\subsection{Baseline Methods}
To compare model performance, we introduce the following learning-based planning baselines.

\textbf{Vanilla Imitation Learning (IL)}: 
A simplified version of our model that directly outputs the planned trajectory of the AV without explicitly reasoning other agents' future trajectories. The plan is only a single-modal trajectory. The original encoder part of our model is utilized, but only one decoder layer with the ego agent's historical encoding as the query is used to decode the AV's plan.

\textbf{Deep Imitative Model (DIM)}: 
A probabilistic planning method that aims to generate expert-like future trajectories $q\left(\mathbf{S}_{1:T}| \phi \right) = \prod_{t=1}^T q\left(\mathbf{S}_{t}| \mathbf{S}_{1:t-1}, \phi \right)$ given the AV’s observations $\phi$. We follow the original open-source DIM implementation and use the rasterized scene image $\mathbb{R}^{200\times200\times3}$ and the AV's historical states $\mathbb{R}^{11\times5}$ as the observation. We use a CNN to encode the scene image and an RNN to encode the agent's historical states. The AV's future state is decoded (as a multivariate Gaussian distribution) in an autoregressive manner. In testing, DIM requires a specific goal $\mathcal{G}$ to direct the agent to the goal, and a gradient-based planner maximizes the learned imitation prior $\log q\left(\mathbf{S} | \phi \right)$ and the test-time goal likelihood $\log p(\mathcal{G} | \mathbf{S}, \phi)$.

\textbf{Robust Imitative Planning (RIP)}: 
An epistemic uncertainty-aware planning method that is developed upon DIM and shows good performance in conducting robust planning in out-of-distribution (OOD) scenarios. Specifically, we employ the original open-source implementation and choose the worst-case model that has the worst likelihood $\min_d \log q\left(\mathbf{S}_{1:T} | \phi \right)$ among $d=6$ trained DIM models and improve it with a gradient-based planner.

\textbf{Conservative Q-Learning (CQL)}: 
A widely-used offline reinforcement learning algorithm that learns to make decisions from offline datasets. We implement the CQL method with the d3rlpy offline RL library\footnote{\url{https://github.com/takuseno/d3rlpy}}. The RL agent takes the same state inputs as the DIM method and outputs the target pose of the next step $\left(\Delta x, \Delta y, \Delta \theta \right)$ relative to the agent's current position. The reward function is the distance traveled per step plus an extra reward for reaching the goal, \ie, $r_t = \Delta d_t + 10 \times \mathbbm{1} \left( d(s_t, s_{goal}) < 1 \right)$. Since the dataset only contains perfect driving data, no collision penalty is included in the reward function.

\textbf{Differentiable Integrated Prediction and Planning (DIPP)}: 
A joint prediction and planning method that uses a differentiable motion planner to optimize the trajectory according to the prediction result. We adopt the original open-source implementation and the same state input setting. We increase the historical horizon to 20 and the number of prediction modalities from 3 to 6. In open-loop testing, we utilize the results from the DIPP prediction network without trajectory planning (refinement).

\textbf{MultiPath++}: 
A high-performing motion prediction model that is based on the context-aware fusion of heterogeneous scene elements and learnable latent anchor embeddings. We utilize the open-source implementation of MultiPath++\footnote{\url{https://github.com/stepankonev/waymo-motion-prediction-challenge-2022-multipath-plus-plus}} that achieved state-of-the-art prediction accuracy on the WOMD motion prediction benchmark. We train the model to predict 6 possible trajectories and corresponding scores for the ego agent using the same dataset. In open-loop testing, only the most-likely trajectory will be used as the plan for the AV.

\textbf{Motion Transformer (MTR)-e2e}: 
A state-of-the-art prediction model that occupies the first place on the WOMD motion prediction leaderboard. We follow the original open-source implementation of the context encoder and MTR decoder. However, we modified the decoder to use an end-to-end variant of MTR that is better suited for the open-loop planning task. Specifically, only 6 learnable motion query pairs are used to decode 6 possible trajectories and scores. The same dataset is used to train the MTR-e2e model, and the data is processed according to the MTR context inputs.

\subsection{Refinement Planner}
\textbf{Inverse dynamic model}.
To convert the initial planned trajectory to control actions $\{a_t, \delta_t\}$ (\ie, acceleration and yaw rate), we utilize the following inverse dynamic model.
\begin{equation}
\begin{split}
\Phi^{-1}: v_t &= \frac{\Delta p}{\Delta t} = \frac{\parallel p_{t+1} - p_{t} \parallel }{\Delta t}, \\
          \theta_t &= \arctan \frac{\Delta p_y}{\Delta p_x}, \\
          a_t &= \frac{v_{t+1} - v_{t}}{\Delta t}, \\
          \delta_t &= \frac{\theta_{t+1} - \theta_{t}}{\Delta t},
\end{split}
\end{equation}
where $p_t$ is a predicted coordinate in the trajectory, and $\Delta t$ is the time interval.

\textbf{Dynamic model}.
To derive the coordinate and heading $\{{p_x}_{t}, {p_y}_{t}, \theta\}$ from control actions, we adopt the following differentiable dynamic model.
\begin{equation}
\begin{split}
\Phi: v_{t+1} &= a_t \Delta t + v_t, \\
     \theta_{t+1} &= \delta_t \Delta t + \theta_t, \\
     {p_x}_{t+1} &= v_t \cos \theta_t {\Delta t} + {p_x}_{t}, \\
     {p_y}_{t+1} &= v_t \sin \theta_t {\Delta t} + {p_y}_{t}.
\end{split}
\end{equation}

\textbf{Motion planner}.
We use a differentiable motion planner proposed in DIPP to plan the trajectory for the AV. The planner takes as input the initial control action sequence derived from the planned trajectory given by our model. We formulate each planning cost term $c_i$ as a squared vector-valued residual, and the motion planner aims to solve the following nonlinear least squares problem:
\begin{equation}
\mathbf{u}^{*} = \arg \min_\mathbf{u} \frac{1}{2} \sum_i \parallel \omega_i c_i(\mathbf{u})  \parallel^2, 
\end{equation}
where $\mathbf{u}$ is the sequence of control actions, and $\omega_i$ is the weight for cost $c_i$.

We consider a variety of cost terms as proposed in DIPP, including travel speed, control effort (acceleration and yaw rate), ride comfort (jerk and change of yaw rate), distance to the reference line, heading difference, as well as the cost of violating traffic light. Most importantly, the safety cost takes all other agents' predicted states into consideration and avoids collision with them, as illustrated in DIPP.

We use the Gauss-Newton method to solve the optimization problem. The maximum number of iterations is 30, and the step size is 0.3. We use the best solution during the iteration process as the final plan to execute.

\textbf{Learning cost function weights}.
Since the motion planner is differentiable, we can learn the weights of the cost terms by differentiating through the optimizer. We use the imitation learning loss below (average displacement error and final displacement error) to learn the cost weights, as well as minimize the sum of cost values. We set the maximum number of iterations to 3 and the step size to 0.5 in the motion planner. We use the Adam optimizer with a learning rate of 5e-4 to train the cost function weights; the batch size is 32 and the total number of training steps is 10,000.
\begin{equation}
\mathcal{L} = \lambda_1 \sum_t ||\hat s_t - s_t||^2 + \lambda_2 ||\hat s_T - s_T||^2 + \lambda_3 \sum_i ||c_i||^2,
\end{equation}
where $\lambda_1 = 1, \lambda_2 = 0.5, \lambda_3 = 0.001$ are the weights.

\subsection{GameFormer Planner}
To validate our model's performance on the nuPlan benchmark\footnote{\url{https://eval.ai/web/challenges/challenge-page/1856/overview}}, we have developed a comprehensive planning framework to handle the realistic driving scenarios in nuPlan. The planning process comprises the following steps: 1) feature processing: relevant data from the observation buffer and map API undergoes preprocessing to extract input features for the prediction model; 2) path planning: candidate route paths for the ego vehicle are computed, from which the optimal path is selected as the reference path; 3) model query: the prediction model is queried to generate an initial plan for the ego vehicle and predict the trajectories of surrounding agents; and 4) trajectory refinement: a nonlinear optimizer is employed to refine the ego vehicle’s trajectory on the reference path and produce the final plan. For computational efficiency, we use a compact version of the GameFormer model, configuring it with 3 encoding layers and 3 decoding layers (1 initial decoding layer and 2 interaction decoding layers). Additionally, we introduce an extra decoding layer after the last interaction decoding layer to separately generate the ego vehicle’s plan. The ego plan is then projected onto the reference path as an initialization of the refinement planner. The output of the GameFormer model consists of multimodal trajectories for the surrounding agents. For each neighboring agent, we select the trajectory with the highest probability and project it onto the reference path using the Frenet transformation, subsequently calculating spatiotemporal path occupancy. A comprehensive description of the planning framework can be found in this dedicated report\footnote{\url{https://opendrivelab.com/e2ead/AD23Challenge/Track_4_AID.pdf}}.

\section{Additional Quantitative Results}
\subsection{Interaction Prediction}
Table \ref{taba} displays the per-category performance of our models on the WOMD interaction prediction benchmark, in comparison with the MTR model. The GameFormer joint prediction model exhibits the lowest minFDE across all object categories, indicating the advantages of our model and joint training of interaction patterns. Our GameFormer model surpasses MTR in the cyclist category and achieves comparable performance to MTR in other categories, though with a much simpler structure than MTR.

\begin{table}[htp]
\caption{Per-class performance of interaction prediction on the WOMD interaction prediction benchmark}
\centering
\resizebox{\linewidth}{!}{%
\begin{tabular}{@{}llllll@{}}
\toprule
Class                       & Model     & minADE ($\downarrow$) & minFDE ($\downarrow$) & Miss rate ($\downarrow$) & mAP ($\uparrow$) \\ \midrule
\multirow{3}{*}{Vehicle}    & MTR       & \textbf{0.9793}       &  2.2157               & 0.3833                   & \textbf{0.2977} \\
                            & GF (J)    & 0.9822                &  \textbf{2.0745}      & \textbf{0.3785}           & 0.1856 \\
                            & GF (M)    & 1.0499                &  2.4044               &  0.4321                    & 0.2469 \\ \midrule
\multirow{3}{*}{Pedestrian} & MTR       & \textbf{0.7098}       & 1.5835                & \textbf{0.3973}           &  \textbf{0.2033} \\
                            & GF (J)    & 0.7279                & \textbf{1.4894}       & 0.4272                    &  0.1505 \\
                            & GF (M)    & 0.7978                & 1.8195                & 0.4713                    &  0.1962 \\ \midrule
\multirow{3}{*}{Cyclist}    & MTR       & 1.0652                & 2.3908                & \textbf{0.5428}          & 0.1102\\
                            & GF (J)    & \textbf{1.0383}       & \textbf{2.2480}       & 0.5536                    &  0.0768\\
                            & GF (M)    & 1.0686                & 2.4199                & 0.5765                     &  \textbf{0.1338} \\ \bottomrule
\end{tabular}
}
\vspace{-0.2cm}
\label{taba}
\end{table}

\subsection{nuPlan Benchmark}
Table \ref{tabb} presents a performance comparison between our planner and the DIPP planner. For the benchmark evaluation, we replace the prediction model in the proposed planning framework with the DIPP model and other parts of the framework remain the same. The results show that the GameFormer model still outperforms the DIPP model, as a result of better initial plans for the ego agent and prediction results for other agents.

\begin{table}[htp]
\caption{Comparison with DIPP planner on the nuPlan testing benchmark}
\centering
\resizebox{\linewidth}{!}{%
\begin{tabular}{@{}lcccc@{}}
\toprule
Method & Overall            & OL                & CL non-reactive   & CL reactivate \\ \midrule
DIPP   & 0.7950             &  0.8141           &  0.7853           & 0.7857              \\
Ours   &\textbf{0.8288}     & \textbf{0.8400}   & \textbf{0.8087}   & \textbf{0.8376}   \\ \bottomrule
\end{tabular}
}
\vspace{-0.5cm}
\label{tabb}
\end{table}

\subsection{Abalation Study}
\textbf{Effects of decoding levels on closed-loop planning}.
We investigate the influence of decoding levels on closed-loop planning performance in selected WOMD scenarios, using the success rate (without collision) as the main metric. We also report the inference time of the prediction network (without the refine motion planner) in closed-loop planning, which is executed on an NVIDIA RTX 3080 GPU. The results in Table \ref{tabf} reveal that increasing the decoding layers could potentially lead to a higher success rate, and even adding a single layer of interaction modeling can bring significant improvement compared to level-$0$. In closed-loop testing, the success rate reaches a plateau at a decoding level of 2, while the computation time continues to increase. Therefore, using two reasoning levels in our model may offer a favorable balance between performance and efficiency in practical applications.

\begin{table}[htp]
\centering
\caption{Effects of decoding levels on closed-loop planning}
\small
\begin{tabular}{@{}l|cc@{}}
\toprule
Level           & Success rate (\%)     & Inference time (ms)      \\ \midrule
0               &  89.5                 & 31.8 \\ 
1               &  92.25                & 44.1 \\
2               &  94                   & 56.7 \\
3               &  94.5                 & 66.5 \\ 
4               &  94.5                 & 79.2 \\ \bottomrule
\end{tabular}%
\label{tabf}
\vspace{-0.1cm}
\end{table}

\section{Additional Qualitative Results}
\subsection{Interaction Prediction}
\begin{figure*}[htp]
    \centering
    \includegraphics[width=\linewidth]{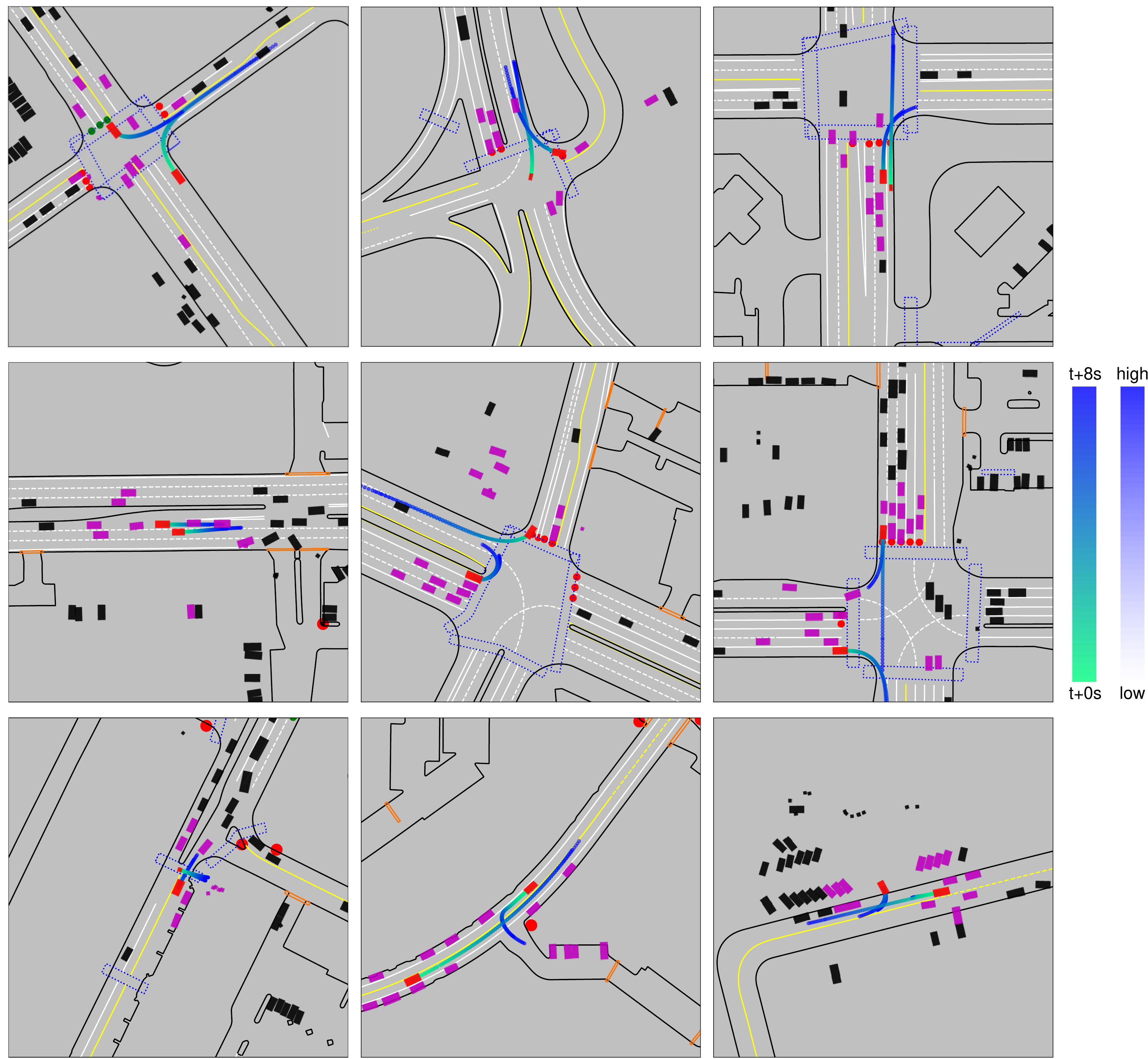}
    \caption{Additional qualitative results of interaction prediction. The red boxes are interacting agents to predict, and the magenta boxes are background neighboring agents. Six joint trajectories of the two interacting agents are predicted.}
    \label{fig6}
\end{figure*}

Fig. \ref{fig6} presents additional qualitative results of our GameFormer framework in the interaction prediction task, showcasing the ability of our method to handle a variety of interaction pairs and complex urban driving scenarios.

\subsection{Level-$k$ Prediction}
\begin{figure*}[htp]
    \centering
    \includegraphics[width=\linewidth]{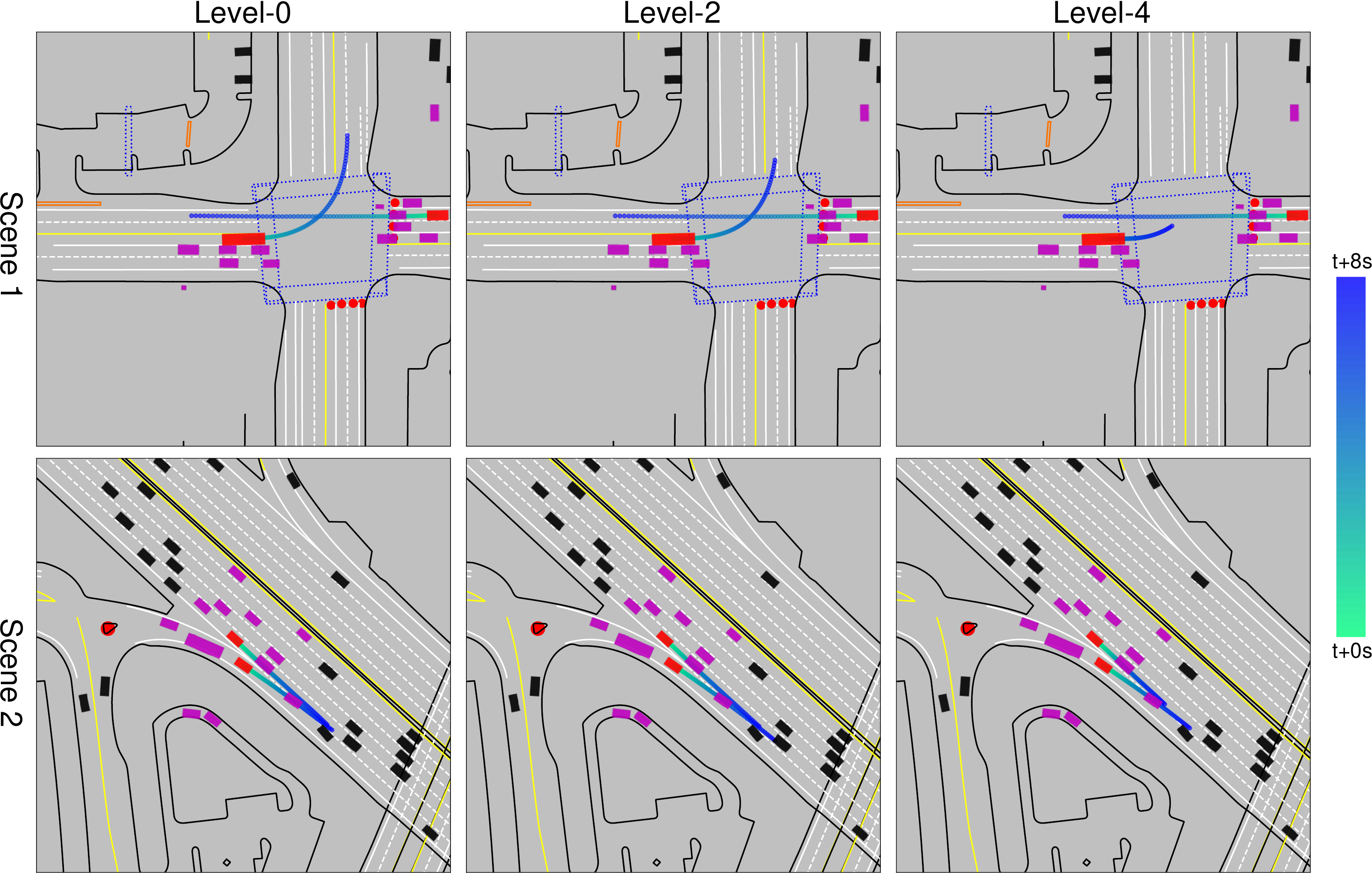}
    \caption{Prediction results of the two interacting agents at different reasoning levels. Only the most-likely joint trajectories of the target agents are displayed for clarity.}
    \label{fig7}
\end{figure*}

Fig. \ref{fig7} illustrates the most-likely joint trajectories of the target agents at different interaction levels. The results demonstrate that our proposed model is capable of refining the prediction results in the iterated interaction process. At level-$0$, the predictions for target agents appear more independent, potentially leading to trajectory collisions. However, through iterative refinement, our model can generate consistent and human-like trajectories at a higher interaction level.

\subsection{Open-loop Planning}
\begin{figure*}[htp]
    \centering
    \includegraphics[width=\linewidth]{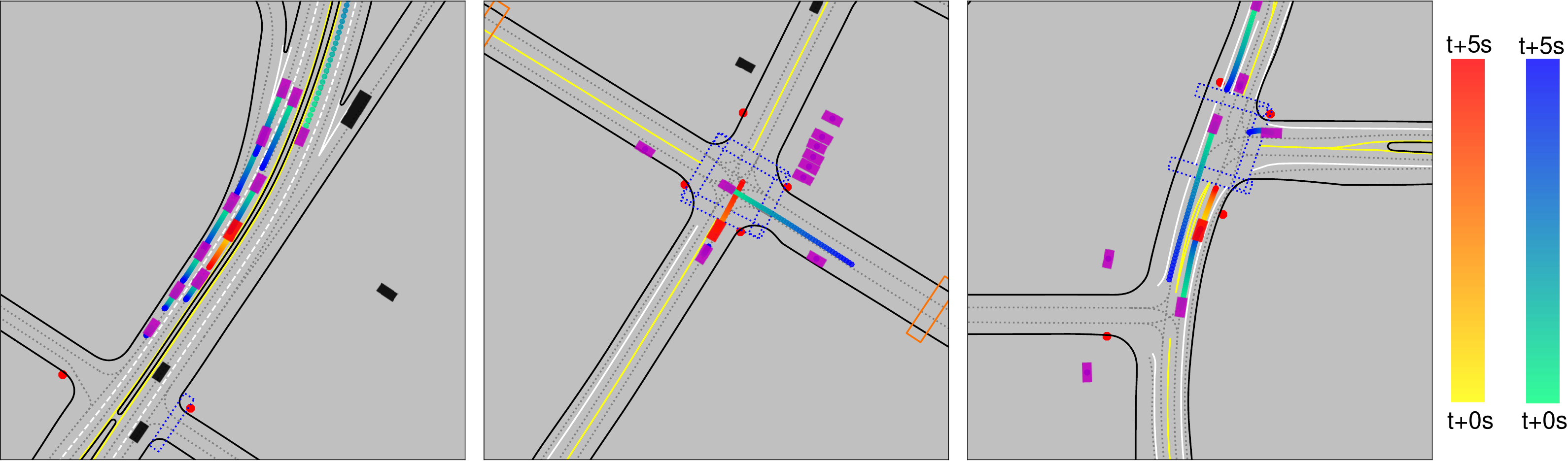}
    \caption{Additional qualitative results of open-loop planning. The red box is the AV and the magenta boxes are its neighboring agents; the red trajectory is the plan of the AV and the blue ones are the predictions of neighboring agents.}
    \label{fig8}
\end{figure*}

Fig. \ref{fig8} provides additional qualitative results of our model in the open-loop planning task, which show the ability of our model to jointly plan the trajectory of the AV and predict the behaviors of neighboring agents.

\subsection{Closed-loop Planning}
We visualize the closed-loop planning performance of our method through videos available on the \href{https://mczhi.github.io/GameFormer/}{project website}, including interactive urban driving scenarios from both the WOMD and nuPlan datasets.

\end{document}